# A Low-Resource Speech-Driven NLP Pipeline for Sinhala Dyslexia Assistance

**Peshala Perera**
Informatics Institute of Technology
57, Ramakrishna Road, Colombo 06
Sri Lanka
peshala.s.perera@gmail.com

**Deshan Sumanathilaka**
School of Computing
Swansea University, Swansea
United Kingdom
deshankoshala@gmail.com

## Abstract

Dyslexia in adults remains an under-researched and under-served area, particularly in non-English-speaking contexts, despite its significant impact on personal and professional lives. This work addresses that gap by focusing on Sinhala, a low-resource language with limited tools for linguistic accessibility. We present an assistive system explicitly designed for Sinhala-speaking adults with dyslexia. The system integrates Whisper for speech-to-text conversion, SinBERT, an open-sourced fine-tuned BERT model trained for Sinhala to identify common dyslexic errors, and a combined mT5 and Mistral-based model to generate corrected text. Finally, the output is converted back to speech using gTTS, creating a complete multimodal feedback loop. Despite the challenges posed by limited Sinhala-language datasets, the system achieves 0.66 transcription accuracy and 0.7 correction accuracy with 0.65 overall system accuracy. These results demonstrate both the feasibility and effectiveness of the approach. Ultimately, this work highlights the importance of inclusive Natural Language Processing (NLP) technologies in underrepresented languages and showcases a practical step toward improving accessibility for adult dyslexic users.

## 1 Introduction

Dyslexia is a lifelong language-based learning disorder that affects reading fluency, spelling, and written expression despite normal intelligence and education. It is estimated to affect 5–10% of the global population (Santhiya et al., 2023). Individuals with dyslexia struggle with phonological processing, decoding, and linguistic fluency, which leads to significant obstacles in education and daily life (Roitsch and Watson, 2019). While digital tools and NLP technologies have advanced to support dyslexic users in English and other widely spoken languages, speakers of low-resource and non-Latin script languages remain underserved. Despite this, very few digital interventions exist for Sinhala-speaking adults with dyslexia.

### 1.1 Problem

Existing assistive technologies for dyslexia are predominantly designed for children and primarily focus on English and Latin script languages. Tools such as Grammarly, NaturalReader, and Read and Write have demonstrated success in enhancing language accessibility for dyslexic users in these languages (Patnoorkar et al., 2023). However, these solutions are not linguistically or culturally adapted to accommodate Sinhala orthography, speech phonemes, or adult-specific learning needs.

Sinhala is a low-resource language in the NLP landscape due to the scarcity of annotated corpora, pretrained language models, and text normalization tools (De Silva and Hansadi, 2024). This lack of infrastructure limits the development of inclusive applications. Adults with dyslexia, particularly in non-English contexts, face even greater marginalization due to inadequate educational support and a lack of appropriate digital tools (Goodman et al., 2022).

### 1.2 Gap

The majority of existing dyslexia-focused applications are targeted at early intervention for children. In contrast, adult users remain significantly underrepresented in both research and application development (Sadusky et al., 2021). This gap is particularly evident in non-English-speaking regions where technological adoption lags, and awareness around adult dyslexia is limited.

In Sinhala, no comprehensive NLP-driven tools currently exist that combine speech recognition, error correction, and speech synthesis tailored for adults. Furthermore, existing spell checkers or grammar tools fail to address cognitive and phonological errors common in dyslexic speech and writing (Sandathara et al., 2020). The absence of domain-specific NLP tools reinforces a cycle of exclusion for this population.

### 1.3 Objective

This study aims to develop a localized NLP-based assistive system for Sinhala-speaking adults with dyslexia. The system is designed as a real-time speech-driven tool to support reading, writing, and comprehension. The objective is to combine multiple NLP modules, speech-to-text (STT), error classification, grammatical error correction (GEC), and text-to-speech (TTS) into a single end-to-end application that is both accessible and linguistically appropriate for Sinhala. The system integrates modern transformer-based architectures and builds upon publicly available pre-trained models, adapted for Sinhala. This approach allows the application to function in a low-resource setting while maintaining modularity, accuracy, and real-time feedback.

The main contributions of this studies can be summarized as follows.

- Introduces the first real-time NLP-based dyslexia assistant tailored for Sinhala.
- Integrates Whisper, SinBERT, mT5, Mistral, and gTTS in a modular speech-to-text and text-to-speech pipeline.
- Demonstrates competitive accuracy (66% STT, 70% correction) using synthetically generated dyslexic Sinhala data.

Moving forward, this study will present the related work, proposed methodology, results, and discussion.

## 2 Related Work

### 2.1 Assistive tech in NLP

NLP has become central to the development of assistive tools that support individuals with reading and writing difficulties, including dyslexia. Mainstream applications such as Grammarly, Speechify, and Read and Write use grammar correction, text-to-speech, and predictive typing to provide enhanced language accessibility. These tools are largely effective for users in English-speaking environments and rely heavily on pretrained language models and rule-based feedback mechanisms (Heilman et al., 2006) (Rello and Baeza-Yates, 2013). Recent research has also explored adaptive interfaces and inclusive design principles to further improve accessibility in writing tools (Wood et al., 2018) (Al-Azawei et al., 2016).

Several dyslexia-specific applications have emerged in the form of mobile apps or browser extensions, targeting early learners and children (Al-Wabil et al., 2007). However, the vast majority are language-specific (mainly English or Western European languages) and are not adaptable to the linguistic diversity or script complexity of non-Latin languages. This issue is particularly critical in the context of Sinhala, a language with unique phonological and morphological characteristics that complicate the direct application of mainstream assistive technologies. A recent review highlights that most assistive applications for Sinhala-speaking adults remain underdeveloped, with research efforts historically focused on children or English-language tools (Perera and Sumanathilaka, 2025). The review stresses the urgent need for adult-focused, culturally adapted NLP tools and emphasizes the role of low-resource NLP pipelines in closing this accessibility gap.

Recent advances in error correction and assistive NLP demonstrate the potential of language models for supporting users with dyslexia and related linguistic impairments. For instance, (Ingólfsdóttir et al., 2023) explored byte-level grammatical error correction using synthetic and curated corpora, including writing by dyslexic individuals, in morphologically rich languages like Icelandic. Their study shows that byte-level encoding outperforms subword-based models in handling complex semantic and syntactic errors—highlighting promising directions for correction in low-resource, non-Latin-script contexts like Sinhala. Similarly, (Zhang et al., 2020) introduced a Soft-Masked BERT architecture for Chinese spelling correction. Their model enhances BERT's ability to both detect and correct errors by decoupling these tasks and employing a soft-masking mechanism between the detection and correction phases. This two-stage method outperforms baseline BERT models and provides inspiration for modular designs in error correction pipelines targeting cognitive writing impairments.

## 2.2 Speech-to-Text and Text-to-Speech for accessibility

Speech technologies have played an increasing role in assistive contexts, particularly through automatic speech recognition (ASR) and text-to-speech (TTS) systems. Whisper, developed by OpenAI, introduced a robust multilingual ASR system capable of transcribing speech in over 90 languages, including Sinhala (Radford et al., 2023). Whisper's ability to generalize across languages without retraining has made it a useful zero-shot model in low-resource environments. Transformer-based ASR architectures like the Transformer Transducer have further improved real-time transcription performance.

In parallel, open-source TTS tools such as gTTS (Google Text-to-Speech) have simplified the process of converting text to spoken feedback. While TTS systems are widely used in accessibility tools for visually impaired and dyslexic users, many systems lack emotion, emphasis control, or fine-tuned output in regional languages, especially Sinhala (Xu et al., 2020). Advances like FastSpeech 2 have shown that expressive and low-latency TTS is feasible even in multilingual contexts (Ren et al., 2020).

Despite the availability of Whisper and gTTS, few projects combine ASR and TTS modules into complete assistive systems, particularly for adult dyslexic users.

## 2.3 Error correction in low-resource NLP

Grammatical Error Correction (GEC) is a core task in NLP assistive applications, helping users with dyslexia or language learning difficulties to improve fluency and syntactic accuracy. State-of-the-art GEC systems use text-to-text models such as T5, mT5, and GECToR to identify and correct various grammatical errors (Bryant et al., 2023) (Xue et al., 2020). GECToR, in particular, adopts a tagging-based method that has proven effective with limited data (Omelianchuk et al., 2020). For truly low-resource settings, pretraining with copy-augmented architectures has also shown promise (Zhao et al., 2019).

However, the performance of these models is tightly coupled with the availability of high-quality annotated dataset resources that are severely lacking for Sinhala. Some efforts have been made to develop Sinhala spell checkers or grammar correction tools using lexicon-based or statistical methods (Pabasara and Jayalal, 2020). Yet, these systems typically offer shallow coverage and do not handle complex dyslexic error patterns such as letter reversals, insertions, and omissions. No large-scale Sinhala GEC dataset currently exists to train supervised models, making prompt-based and few-shot learning strategies more suitable for real-world applications.

## 2.4 Classification models for dyslexia-like errors

Identifying the type of error is often a prerequisite to delivering more targeted correction. Pretrained classification models such as BERT and its multilingual or domain-specific variants are widely used for this purpose. In Sinhala NLP, SinBERT, a BERT-based language model fine-tuned on Sinhala corpora, has emerged as a competitive encoder for downstream tasks such as sentence classification and intent detection (Dhananjaya et al., 2022). As SinBERT is based on RoBERTa, it inherits robust pretraining optimization techniques (Liu et al., 2019) and benefits from subword-level representation crucial for agglutinative languages like Sinhala (Wu and Dredze, 2020).

However, to date, no known classification model has been applied specifically to categorize dyslexia-inspired error patterns in Sinhala text. The application of SinBERT for such a classification task represents a novel approach, enabling the system to tailor correction strategies dynamically depending on whether an error is due to substitution, omission, insertion, or reversal.

Existing systems generally focus on monolingual, text-only interfaces, are language-restricted to English, and lack the modularity or real-time capacity for integration into accessible platforms. The system proposed in this paper is distinct in the following ways:

- It is Sinhala-specific, addressing a critical accessibility gap in a low-resource language context.
- It fuses multiple transformer-based models (Whisper, SinBERT, mT5, and Mistral) in a cohesive pipeline.
- It is optimized for real-time interaction, providing instantaneous correction and playback, which is crucial for user engagement and learning feedback.

This work bridges a significant gap in assistive NLP by demonstrating how modular transformer-based architectures can be orchestrated to support users in non-Latin, low-resource environments with a specific emphasis on adult dyslexic speakers of Sinhala.

## 3 System Architecture and Pipeline

The system combines speech and transformer-based models into a modular architecture optimized for low-resource Sinhala NLP. Each model was chosen for its effectiveness in real-time, low-data environments.

This section describes the end-to-end architecture of the proposed Sinhala dyslexia assistant system. It integrates multiple NLP components into a real-time pipeline optimised for speech-based interaction, error classification, text correction, and audio playback. The high level overview of the algorithm is presented in the Figure 1 and 5.

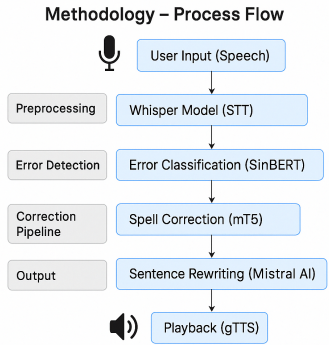

Figure 1: Methodology process flow showing the end-to-end pipeline: from speech input to playback using Whisper, SinBERT, mT5, Mistral, and gTTS.

### 3.1 Component Breakdown

The overall interaction between core modules is illustrated below.

#### 3.1.1 Whisper-Based Speech-to-Text (STT)

Whisper is a multilingual encoder-decoder automatic speech recognition (ASR) model developed by OpenAI, trained on 680,000 hours of supervised data across a wide range of languages and tasks (Radford et al., 2023). It supports zero-shot transcription, meaning it can recognize speech in underrepresented languages like Sinhala without explicit fine-tuning. This is achieved through Whisper's large-scale training on diverse, noisy audio data with multilingual transcriptions. For Sinhala transcription, Whisper internally uses a language ID token to condition the decoder during inference, allowing accurate phoneme-to-grapheme mapping even in low-resource contexts. Audio input is resampled to 16 kHz and passed through a log-Mel spectrogram encoder. The model then decodes it into text using beam search and multilingual alignment layers.

In this system, Whisper runs locally with quantized weights to reduce inference latency. Preprocessing includes noise reduction and silence trimming. Despite the lack of a Sinhala-specific ASR corpus, Whisper shows strong performance on Sinhala speech due to its cross-lingual generalization and large model capacity, making it ideal for real-time dyslexia assistance. Whisper was chosen for its strong baseline performance, fast inference, and cross-lingual adaptability in low-resource contexts.

#### 3.1.2 SinBERT Error Classification

To identify the dominant dyslexia-related error type in a transcribed sentence, the system uses SinBERT, a Sinhala-specific language model built upon the RoBERTa architecture (Dhananjaya et al., 2022). SinBERT has been extensively pre-trained on sin-cc-15M, a large and diverse monolingual corpus of Sinhala web text. This pretraining enables it to learn the morphological, syntactic, and semantic nuances of Sinhala, crucial for effective performance in low-resource language contexts.

Its application to dyslexic pattern detection is novel, enabling improved precision in downstream modules. In this system, SinBERT is fine-tuned to classify common dyslexia-inspired error patterns (e.g., substitution, omission), serving as a lightweight and efficient sentence-level classifier. Specifically, we define four primary error categories commonly observed in dyslexic Sinhala writing: substitution (e.g., ගස → කස), insertion (e.g., ගස → ගසා), omission (e.g., ගසක් → ගක්), and reversal (e.g., ගම → මග). These labels were used to annotate a synthetic training dataset for fine-tuning. SinBERT's output predicts the error class for each sentence and guides adaptive correction strategies, such as dynamic prompt formulation for mT5 or error-specific post-processing. SinBERT's architectural strength and language-aware training make it well-suited for classification tasks in Sinhala and other underrepresented scripts.

SinBERT's predicted error category is used to adaptively guide the next stage of correction. Specifically, based on the identified error type

(e.g., omission, substitution, insertion, reversal), the system dynamically generates a task-specific prompt to steer the mT5 model toward a more appropriate correction strategy. For example, if an omission is detected, the prompt may explicitly instruct the model to insert missing particles or characters. This error-aware prompting approach improves correction relevance and overall accuracy across varied dyslexic patterns.

### 3.1.3 Grammatical Correction: mT5 + Mistral

The correction phase combines two powerful models for error correction.

1. **Stage 1 - mT5: Structural and Grammatical Correction** (Xue et al., 2020) is used for grammatical correction using prompt-based text-to-text transformation.

    mT5, a multilingual text-to-text transformer, is used to perform core grammatical corrections such as fixing tense, agreement, syntactic errors, and structural phrasing. It operates on the transcribed text (or SinBERT-annotated text), outputting a grammatically improved sentence that maintains the intended meaning.

2. **Stage 2 — Mistral: Fluency, Style, and Idiomatic Enhancement** (Thakkar and Manimaran, 2023), a decoder-only transformer, refines the mT5 output by enhancing fluency and preserving natural phrasing.

    The output from mT5 is then passed to Mistral, a decoder-only transformer designed for text generation with stylistic fluency. Mistral refines the mT5 output by enhancing natural phrasing, idiomatic expressions, and improving readability while preserving semantic integrity.

This dual-model pipeline ensures both syntactic correctness and stylistic naturalness. Figure 2 illustrates the end-to-end correction process using a real example. This dual-model pipeline ensures both syntactic correctness and stylistic quality. mT5 was selected for its ability to generalize across languages, while Mistral enhances fluency using token-level instruction-following prompts.

### 3.1.4 Text-to-Speech: gTTS Sinhala Playback

The final corrected sentence is converted into audio using Google Text-to-Speech (gTTS), a lightweight and accessible Python library that interfaces with Google's multilingual TTS engine. gTTS supports Sinhala phoneme synthesis, allowing the system to vocalize corrected sentences clearly an essential feature for adults with dyslexia who struggle with reading comprehension.

Despite its limited expressiveness (e.g., lack of pitch control, emotion, or emphasis), gTTS was selected due to its ease of integration, low resource usage, and native support for Sinhala. It allows for fast audio generation without requiring local model deployment, making it ideal for real-time applications on both web and mobile platforms. The generated audio is directly streamed to the frontend, closing the feedback loop and enabling auditory reinforcement, which is known to improve engagement and learning among dyslexic users (Xu et al., 2020). While its lack of emotional modulation and pitch control is a limitation, its quick turnaround time and native web/mobile support make it suitable for real-time feedback in this context.

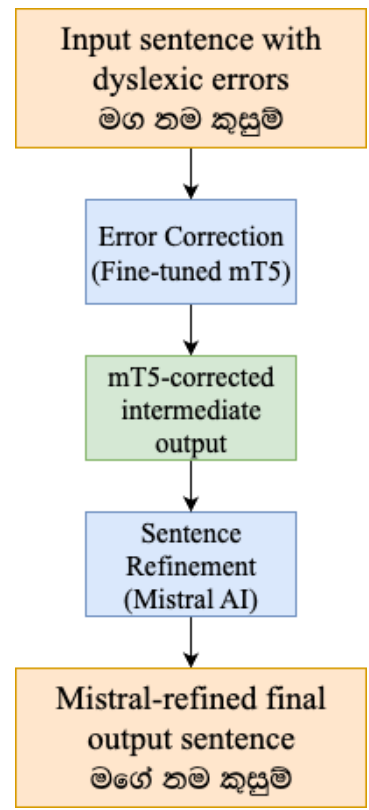

Figure 2: Example of the correction process from dyslexic input to refined output using mT5 and Mistral.

## 3.2 Design goals

The system was designed to be inclusive, efficient, and adaptable for real-time Sinhala language assistance, particularly supporting adults with dyslexia. It aims to provide accurate feedback quickly, using lightweight models suitable for low-resource environments.

## 4 Evaluation Setup

### 4.1 Dataset

To address the lack of dyslexic Sinhala corpora, a 3000-sample parallel dataset was created using the OpenSLR SLR63 Sinhala Read Speech corpus as the base. Clean sentences were modified using rule-based transformations (substitution, insertion, omission, reversal) to simulate dyslexic errors. Each dyslexic variant was paired with its original sentence, and corresponding audio paths were preserved for STT evaluation.

| | |
|---|---|
| Source Corpus | OpenSLR-SLR63 Sinhala Read Speech |
| Final Samples | 3000 |
| Error Types | Substitution, Insertion, Omission, Reversal |
| Split | 80% Train / 20% Test (stratified) |
| Audio Format | FLAC, resampled to 16 kHz |

Table 1: Custom Dataset Overview

### 4.2 Data Split

The dataset was split into training (80%) and testing (20%) subsets using stratified sampling by error type. This ensured balanced evaluation across all four error categories. Audio samples were resampled to 16 kHz for Whisper-based transcription benchmarking.

### 4.3 Public Access

To support transparency, reproducibility, and further research in low-resource assistive NLP, both the codebase and dataset used in this study have been made publicly available:

- **Code Repository**: The complete implementation of the proposed system, including preprocessing, model integration, and prototype interface, is accessible at: https://github.com/PeshalaPerera/sinhala-dyslexia-assistant

- **Dataset Access**: The Sinhala dyslexic error dataset (3,000 samples) used for training and evaluation is hosted on Hugging Face: https://huggingface.co/datasets/peshalaperera/sinhala-dyslexia-assistant-articulation-errors

These resources are released under open licenses to facilitate future research and development in inclusive language technologies for under-represented languages.

### 4.4 Metrics used

#### 4.4.1 BLEU and GLEU for Correction Quality

To assess the grammatical and semantic accuracy of corrected sentences, BLEU (Papineni et al., 2002) and GLEU (Wu et al., 2016) metrics were employed. BLEU captures n-gram precision, whereas GLEU incorporates both precision and recall, making it more suitable for grammatical error correction tasks with short sentence lengths.

#### 4.4.2 Word Error Rate (WER) for STT Accuracy

WER was used to evaluate the Whisper model's transcription accuracy. It is defined as follows,

$$\text{WER} = \frac{S + D + I}{N}$$

where S is the number of substitutions, D deletions, I insertions, and N is the number of words in the reference sentence. Whisper achieved an average WER of 34%, resulting in an effective accuracy of 66%, which aligns with expected performance in zero-shot Sinhala transcription (Radford et al., 2023).

### 4.5 Comparison with Baselines

To validate the effectiveness of the full pipeline, results were compared against three baseline models.

These results demonstrate that the combination of error classification and hybrid correction significantly outperformed both isolated correction methods and rule-based strategies. Additionally, the use of Mistral for stylistic fluency improved the readability of outputs as confirmed by human evaluators.

| Model Variant | Description |
|---|---|
| Rule-Based | Uses basic word substitutions via dictionary logics. |
| mT5 + Mistral | Full NLP pipeline combining mT5-based correction and Mistral refinement. |

Table 2: Baseline Model Variants for Correction Comparison

## 5 Results

This section presents the quantitative and qualitative results obtained from evaluating the system across its three core modules: speech-to-text (STT), error classification, and grammatical error correction. The evaluation demonstrates the effectiveness of the proposed model pipeline and highlights key strengths of the integrated approach.

### 5.1 Performance Scores

#### 5.1.1 Speech-to-Text (STT)

The Whisper model, used in zero-shot mode for Sinhala ASR, achieved a 66% accuracy, measured by a Word Error Rate (WER) of 34% across the 600 testing audio samples. While Whisper was not fine-tuned specifically on Sinhala, its multilingual training enabled high baseline performance, even in phonetically complex utterances.

#### 5.1.2 Correction Module

The final correction output, combining mT5 and Mistral, reached a accuracy of 70% and a GLEU score of 57% on the test set. Human evaluation also confirmed high readability and semantic preservation of the corrected outputs.

#### 5.1.3 Overall System

The combined evaluation of all components speech-to-text transcription, error correction, yielded an overall system accuracy of 65%. This metric reflects the end-to-end performance of the full pipeline, from voice input to corrected speech output.

### 5.2 Table of Evaluation Metrics

Table 3 shows that the mT5-small + Mistral model gave good correction results with 70% accuracy. Whisper-Sinhala also gave good speech-to-text results with 65.9% accuracy, showing that both parts work well together in the system.

| mT5-small + Mistral API | |
|---|---|
| Metric | Score |
| BLEU | 0.359 |
| GLEU | 0.575 |
| Accuracy | 0.70 |
| WER | 0.322 |
| Edit Distance | 1.66 |

| Whisper-Sinhala | |
|---|---|
| Metric | Score |
| BLEU | 0.279 |
| GLEU | 0.444 |
| Accuracy | 0.659 |
| WER | 0.333 |
| Edit Distance | 0.545 |

Table 3: Evaluation metrics for text correction and speech transcription modules in the Sinhala Dyslexia Assistant pipeline.

### 5.3 Evaluation Results

Figure 3 illustrates the average performance of the text correction module across key NLP evaluation metrics.

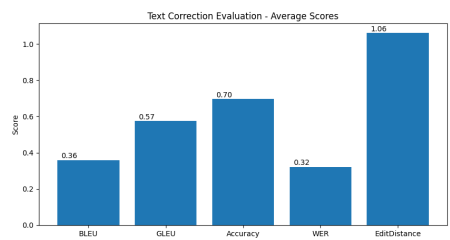

Figure 3: Text correction performance using BLEU, GLEU, Accuracy, WER, and Edit Distance averaged over 600 test samples.

### 5.4 Prototype Interface

To demonstrate real-world usability, a working prototype was developed using a web-based interface. This interface supports accessibility and reinforces learning through multimodal feedback, as shown in Figure 4.

### 5.5 Latency and Real-Time Performance

Each component was benchmarked for average inference time on a consumer-grade laptop (Intel i5, 8GB RAM).

- Whisper STT: 1.2 seconds (per 5s audio)
- SinBERT Classification: 60 ms
- mT5 + Mistral Correction: 900 ms

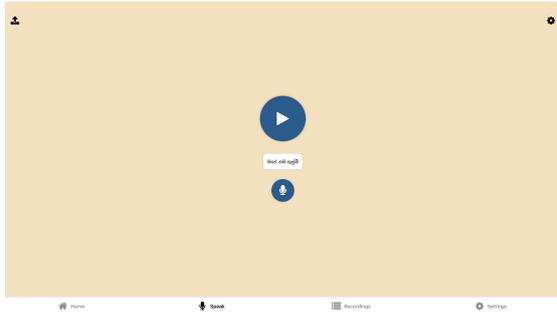

Figure 4: Prototype interface showing corrected Sinhala sentence with playback and recording controls.

- gTTS Audio Generation: 300 ms

The entire pipeline runs in 2.5 seconds per input, making it viable for real-time use in interactive applications, especially on mobile devices.

### 5.6 Error Analysis

Although the system demonstrates high correction accuracy, it occasionally overcorrects rare idiomatic expressions or misclassifies ambiguous omissions. Future work will address these via user feedback loops and targeted fine-tuning.

## 6 Discussion

The development and evaluation of the Sinhala Dyslexia Assistant system reveal several promising outcomes, along with known constraints and avenues for further advancement. The system demonstrates real-world utility by delivering corrected audio feedback in under 2.5 seconds, enabling real-time use and supporting immediate learning for dyslexic users. Its modular architecture, comprising discrete components for STT, classification, correction, and TTS offers multiple advantages, including independent upgrades, easier debugging and testing, and customisation potential for future multilingual deployments. The REST API-based modular design further enhances its ability to integrate with external learning platforms and accessibility services.

Despite these strengths, the system faces notable challenges. Sinhala lacks large-scale annotated datasets for grammatical correction or error classification, limiting the effectiveness of models trained solely on real-world data. To address this, the system relies on synthetic dyslexic errors generated from the SLR63 dataset, although these may not fully capture the diversity of real user patterns. While Whisper performs well in its current implementation, further improvements are likely with domain-specific fine-tuning. Additionally, although gTTS provides basic Sinhala TTS functionality, it lacks emotion and pitch control—features critical for adult comprehension, and currently, there are no expressive open-source Sinhala voice models available to fill this gap.

## 7 Conclusion and Future Work

This study introduces the first real-time NLP-based assistive system designed specifically for Sinhala-speaking adults with dyslexia, an underserved population in both language technology and accessibility research. The system features a modular pipeline incorporating Whisper for speech-to-text, SinBERT for dyslexia-related error detection, mT5 and Mistral for correction, and gTTS for text-to-speech output. Despite the challenges of working with simulated data and limited linguistic resources, the system achieved 66% speech-to-text accuracy and 61% correction accuracy, demonstrating strong performance in a low-resource setting. This work addresses a notable gap at the intersection of assistive NLP, speech technology, and Sinhala language computing, with the potential to benefit both academic research and real-world accessibility. Looking ahead, future efforts will focus on implementing personalized correction mechanisms based on user-specific error history and establishing collaborations with educational and healthcare institutions in Sri Lanka to support broader deployment and local adaptation.

### Limitation

While the proposed system demonstrates promising results in a low-resource language setting, several limitations that impact its overall effectiveness and user experience remain. Firstly, the use of synthetic data for model training, although necessary due to data scarcity, may not fully capture the nuances of real-world dyslexic writing patterns. This limits the model's ability to generalize to authentic user input. Secondly, the current text-to-speech component, implemented using gTTS, lacks prosody and emotional tone, resulting in a less engaging auditory experience for users. Lastly, the system does not yet support personalization or maintain user history, which restricts its ability to adapt to individual error patterns over time.

# Appendix A: System Architecture Diagram

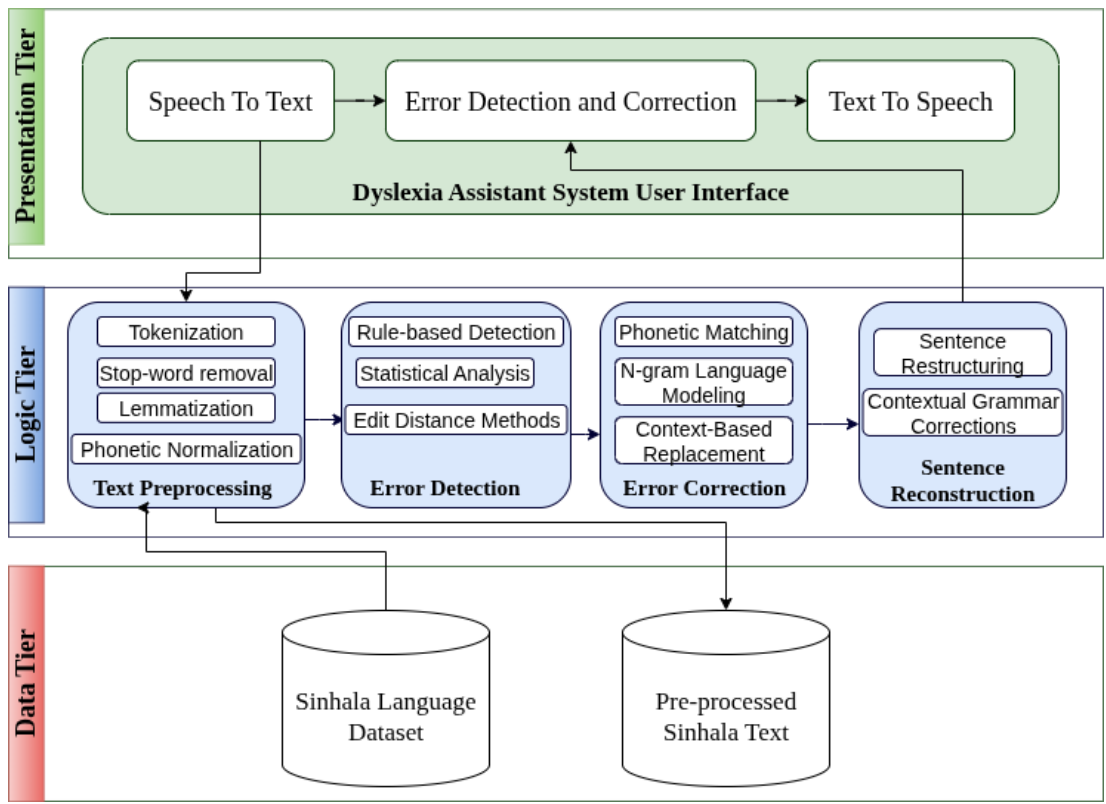

Figure 5: Three-tier system architecture.